\pdfoutput=1

\documentclass[11pt]{article}

\usepackage[]{emnlp2021}

\usepackage{times}
\usepackage{latexsym}

\usepackage[T1]{fontenc}

\usepackage[utf8]{inputenc}

\usepackage{microtype}

\usepackage{booktabs}
\usepackage{graphicx}
\usepackage{amsmath}
\usepackage{multirow}
\usepackage{multicol}
\usepackage{ulem}
\usepackage{capt-of}
\usepackage{varwidth}
\usepackage{xcolor,colortbl}

\definecolor{Gray}{gray}{0.9}

\title{{HRKD}: Hierarchical Relational Knowledge Distillation for Cross-domain Language Model Compression}

\author{Chenhe Dong$^1$, Yaliang Li$^2$, Ying Shen$^{1*}$, Minghui Qiu$^2$\thanks{\ \ Corresponding author.} \\
 $^1$ Sun Yat-sen University $^2$ Alibaba Group\\
 {\tt dongchh@mail2.sysu.edu.cn, sheny76@mail.sysu.edu.cn} \\
 {\tt \{yaliang.li, minghui.qmh\}@alibaba-inc.com}}

\begin{document}

\maketitle

\begin{abstract}
On many natural language processing tasks, large pre-trained language models (PLMs) have shown overwhelming performances compared with traditional neural network methods. Nevertheless, their huge model size and low inference speed have hindered the deployment on resource-limited devices in practice. In this paper, we target to compress PLMs with knowledge distillation, and propose a hierarchical relational knowledge distillation (HRKD) method to capture both hierarchical and domain relational information. Specifically, to enhance the model capability and transferability, we leverage the idea of meta-learning and set up domain-relational graphs to capture the relational information across different domains. And to dynamically select the most representative prototypes for each domain, we propose a hierarchical compare-aggregate mechanism to capture hierarchical relationships. Extensive experiments on public multi-domain datasets demonstrate the superior performance of our HRKD method as well as its strong few-shot learning ability. For reproducibility, we release the code at \url{https://github.com/cheneydon/hrkd}.
\end{abstract}

\section{Introduction}
Large pre-trained language models (PLMs) (e.g., BERT \citep{devlin2019bert}) have demonstrated their outperforming performances on a wide range of NLP tasks, such as machine translation \citep{conneau2019cross-lingual,zhu2020incorporating}, summarization \citep{zhang2019hibert,liu2019text}, and dialogue generation \cite{bao2020plato,zheng2020a-pre-training}. However, their large size and slow inference speed have hindered practical deployments, such as deploying on resource-constrained devices. 

To solve the above problem, many compression techniques for PLMs have been proposed, such as quantization \citep{shen2020q-bert}, weight pruning \citep{michel2019are-sixteen}, and knowledge distillation (KD) \citep{sun2019patient,jiao2020tinybert}. Due to the plug-and-play feasibility of KD, it is the most commonly used method in practice, and we focus on it in this work. The purpose of KD is to transfer knowledge from a larger teacher model to a smaller student model \citep{hinton2015distilling}. Traditional KD methods only leverage single-domain knowledge, i.e., transferring the knowledge of the teacher model to the student model domain by domain. However, as stated in the purpose of transfer learning, the model performance on target domains can be improved by transferring the knowledge from different but related source domains \citep{lu2015transfer}, thus the cross-domain knowledge also plays an important role. In addition, several recent works have also proved the advantage of cross-domain knowledge, and many multi-domain KD methods have been proposed. For example, \citet{peng2020mtss,yang2020model} demonstrate the effectiveness of distilling knowledge from multiple teachers in different domains; \citet{liu2019mkd,liu2019improving} show that jointly distilling the student models of different domains can enhance the performance.

Nevertheless, these methods fail to capture the relational information across different domains and might have poor generalization ability. To enhance the transferability of the multi-domain KD framework, some researchers have recently adopted the idea of meta-learning. Some studies have pointed out that meta-learning can improve the transferability of models between different domains \citep{finn2017model-agnostic,javed2019meta-learning}. For example, Meta-KD \citep{pan2020meta-kd} introduces an instance-specific domain-expertise weighting technique to distill the knowledge from a meta-teacher trained across multiple domains to the student model. However, the Meta-KD framework trains student models in different domains separately, which is inconvenient in real-world applications and might not have enough capability to capture multi-domain correlations.

In this paper, we aim to simultaneously capture the relational information across different domains to make our framework more convenient and effective. Specifically, we set up several domain-relational graphs 
to adequately learn the relations of different domains and generate a set of domain-relational ratios to re-weight each domain during the KD process. Moreover, since different domains might have different preferences of layer prototypes, motivated by the Riesz representation theorem~\citep{hartig1983the-riesz}, we first construct a set of reference prototypes for each domain, which is calculated by a self-attention mechanism to integrate the information of different domains. Then we introduce a hierarchical compare-aggregate mechanism to compare each layer prototype with the corresponding reference prototype and make an aggregation based on their similarities. The aggregated prototypes are finally sent to the corresponding domain-relational graphs. Our framework is referred to as hierarchical relational knowledge distillation (HRKD).

We evaluate the HRKD framework on two multi-domain NLP datasets, including the MNLI dataset \citep{williams2018a-broad-converage-mnli} and the Amazon Reviews dataset \citep{blitzer2007biographies-amazon-review}. Experiments show that our HRKD method can achieve better performance compared with several multi-domain KD methods. We also evaluate our approach under the few-shot learning setting, and it can still achieve better results than the competing baselines.

\section{Method}
\begin{figure*}
\centering
\includegraphics[trim=0 390 120 0, width=0.98\textwidth, clip]{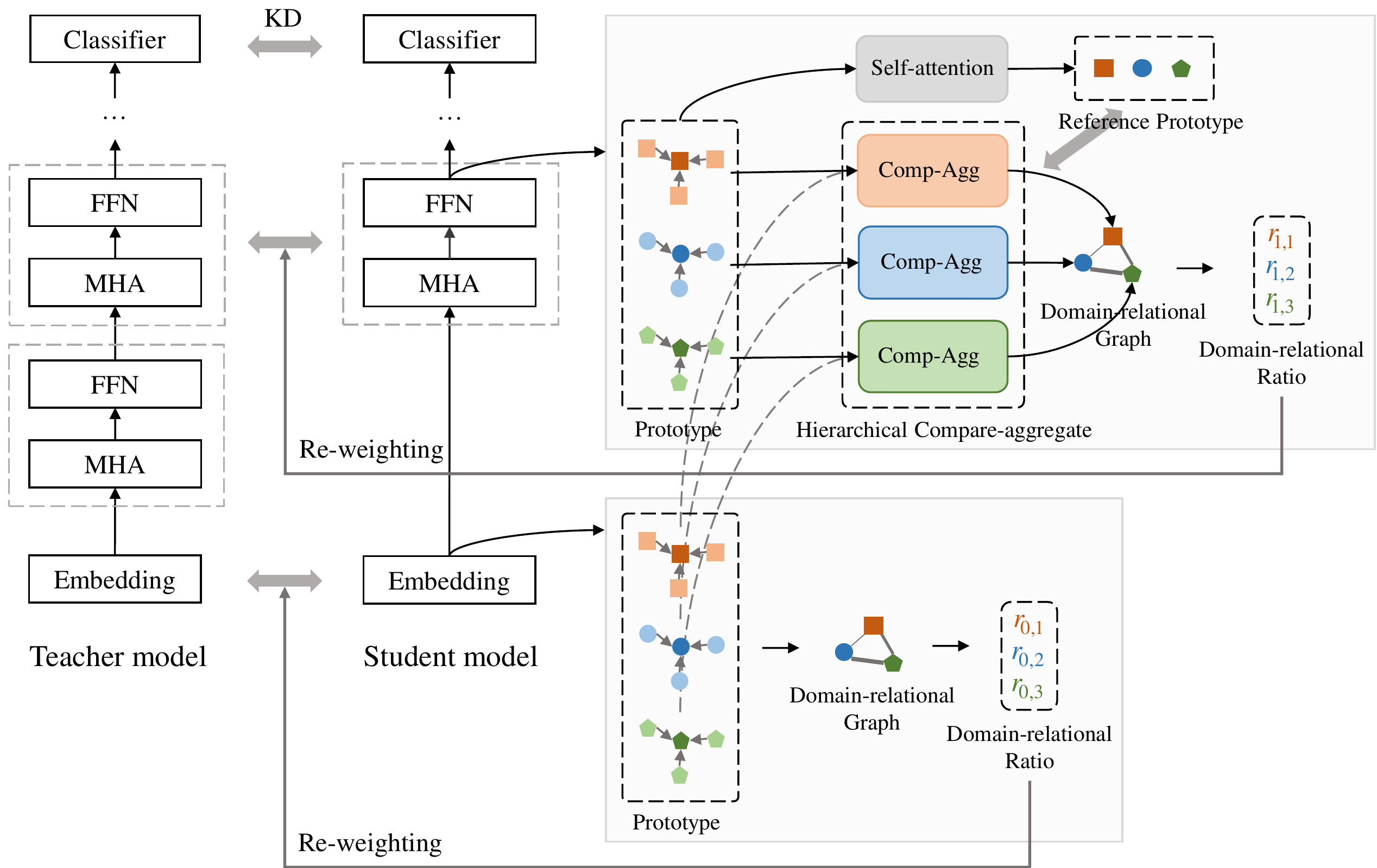}
\caption{An overview of the proposed HRKD method. We use knowledge distillation (KD) to transfer the knowledge from the teacher model to the student model. During KD, we set up several domain-relational graphs to generate domain-relational ratios for re-weighting each domain. We then introduce a hierarchical compare-aggregate mechanism. The prototypes of different layers are dynamically aggregated based on the similarity ratios compared with the corresponding reference prototypes, which are then fed into the domain-relational graphs.}\label{fig: framework}
\end{figure*}

In this section, we detailedly describe the proposed HRKD framework. Our HRKD aims to simultaneously capture the relational information across different domains with both hierarchical and domain meta-knowledges. To achieve this goal, we introduce a hierarchical compare-aggregate mechanism to dynamically identify more representative prototypes for each domain, and construct a set of domain-relational graphs to generate re-weighting KD ratios. The overview of HRKD is shown in Figure \ref{fig: framework}. We first introduce the basic multi-domain KD method in Section \ref{ssec: multi-domain kd}, which is a naive framework lacking the ability of capturing cross-domain relations. Then we describe the domain-relational graph and compare-aggregate mechanism in Section \ref{ssec: domain-relational} and \ref{ssec: compare-aggregate}, respectively, which are the primary modules of our HRKD method to discover the relational information.

\subsection{Multi-domain Knowledge Distillation}
\label{ssec: multi-domain kd}

Similar to \citep{jiao2020tinybert}, we jointly distill the embeddings, attention matrices, transformer layer outputs, and predicted logits between the teacher and student models. Inspired by \citep{liu2019multi-task}, we use a multi-task training strategy to perform multi-domain KD. Specifically, we share the weights of the embedding and transformer layers for all domains while assigning different prediction layers to different domains. Innovatively, we optimize models in different domains simultaneously rather than sequentially.

In detail, the embedding loss $\mathcal{L}_{embd}^d$ and prediction loss $\mathcal{L}_{pred}^d$ of $d$-th domain are formulated as:\begin{align}
\mathcal{L}_{embd}^d &= \hbox{MSE}(\mathbf{E}^S \mathbf{W}^{embd}, \mathbf{E}_d^T), \\
\mathcal{L}_{pred}^d &= \hbox{CE}(\mathbf{z}_d^S / t, \mathbf{z}_d^T / t),
\end{align}where MSE and CE represent the mean square loss and cross-entropy loss, respectively. $\mathbf{E}^S$ and $\mathbf{E}_d^T$ represent the embeddings of student model and teacher model of $d$-th domain, respectively. $\mathbf{z}_d^S$ and $\mathbf{z}_d^T$ represent the predicted logits of student model and teacher model of $d$-th domain, respectively. $\mathbf{W}^{embd}$ is a learnable transformation matrix to align the student embedding dimension that mismatches with the teacher embedding dimension, and $t$ is the temperature factor. 

The attention loss $\mathcal{L}_{attn}^{m, d}$ and transformer layer output loss $\mathcal{L}_{hidn}^{m, d}$ at $m$-th student layer and $d$-th domain are formulated as:\begin{align}
\mathcal{L}_{attn}^{m,d} &= \frac{1}{h} \sum_{i = 1}^h \hbox{MSE}(\mathbf{A}_{i,m}^S, \mathbf{A}_{i,n,d}^T), \\
\mathcal{L}_{hidn}^{m,d} &= \hbox{MSE}(\mathbf{H}_m^S \mathbf{W}_m^{hidn}, \mathbf{H}_{n,d}^T),
\end{align}where $h$ is the number of attention heads, $\mathbf{A}_{i,m}^S$ and $\mathbf{A}_{i,n,d}^T$ are the $i$-th head of attention matrices at $m$-th student layer and its matching $n$-th teacher layer of $d$-th domain, respectively. $\mathbf{H}_m^S$ and $\mathbf{H}_{n,d}^T$ are the transformer layer outputs at $m$-th student layer and $n$-th teacher layer of $d$-th domain, respectively. $\mathbf{W}_m^{hidn}$ is a transformation matrix to align the $m$-th layer of student output dimension that mismatches with the $n$-th layer of teacher output dimension. We use uniform strategy to match the layers between the student and teacher models. 

Finally, the overall KD loss is formulated as: \begin{equation}
\resizebox{\columnwidth}{!}{
$\begin{aligned} 
\mathcal{L}_{total} = \sum_{d=1}^D & \left( {\mathcal{L}_{embd}^d + \sum_{m=1}^M (\mathcal{L}_{attn}^{m,d} + \mathcal{L}_{hidn}^{m,d})} \right. \\ 
&\left. {+ \gamma \mathcal{L}_{pred}^d} \right),
\end{aligned}$}
\end{equation}where $D$ is the total domain number, $M$ is the number of transformer layers in the student model, $\gamma$ is used to control the weight of the prediction loss $\mathcal{L}_{pred}$.

\subsection{Prototype-based Domain-relational Graph}\label{ssec: domain-relational}
Although the basic multi-domain KD method described in Section \ref{ssec: multi-domain kd} can distill the student models across different domains, the relational information between different domains is neglected, which is important for enhancing the model transferability as pointed out by previous studies \cite{finn2017model-agnostic,javed2019meta-learning}. To solve the problem, we attempt to leverage meta-learning to enhance the performance and transferability of our student model. Inspired by the metric-based methods of meta-learning \cite{snell2017prototypical,sung2018learning}, we use prototype representations rather than raw samples to reflect the characteristics of each domain data. This helps to alleviate the negative impact of abnormal samples when there are few training samples (e.g., overfitting) and make the meta-learner easier to learn transferable cross-domain knowledge. Moreover, since we conduct KD over all of the student layers, we calculate different prototypes for different student layers to explicitly distinguish their characteristics. Specifically, the prototype $\mathbf{h}_{m,d}$ of $m$-th layer of the student model at $d$-th domain is calculated by:\begin{equation}\label{eq: h_m^d}
\resizebox{\columnwidth}{!}{
$\mathbf{h}_{m,d} = \left\{
\begin{array}{ll}
\frac{1}{|\mathcal{D}_d| L} \sum_{i=1}^{|\mathcal{D}_d|} \sum_{l=1}^L \mathbf{E}_{i,l}^S, & m = 0 \\
\frac{1}{|\mathcal{D}_d| L} \sum_{i=1}^{|\mathcal{D}_d|} \sum_{l=1}^L \mathbf{H}_{m,i,l}^S, & 1 \leq m \leq M
\end{array}
\right.$}
\end{equation}where $\mathcal{D}_d$ refers to the training set of $d$-th domain, $L$ refers to the sentence length (i.e., number of tokens), $\mathbf{E}_{i,l}^S$ represents $l$-th token of $i$-th sampled student embedding in $\mathcal{D}_d$, and $\mathbf{H}_{m,i,l}^S$ represents the $l$-th token output by the $i$-th sampled student transformer layer of the $m$-th student layer in $\mathcal{D}_d$. In practice, we calculate different prototypes for different batches of training samples.

Afterward, these domain prototypes are leveraged to probe the relations across different domains. Although many multi-domain text mining methods have been proposed recently \cite{wang2020meta,pan2020meta-kd}, they capture the relations separately for each given domain, which might be inconvenient and time-consuming in practice. Meanwhile, the learning process is not effective enough since the other domains cannot learn from each other when optimizing a specific domain. To solve this problem, we aim to simultaneously discover the cross-domain relations to make our framework more convenient and effective. To achieve the goal, we propose to use the graph attention network (GAT) \cite{velickovic2018graph} to process the prototypes of all domains at the same time. To utilize GAT, each node in the graph represents a domain prototype, and each edge weight represents the similarity of the connected two prototypes. In this way, the relations across different domains can be captured simultaneously. In detail, we set up a two-layer \textit{domain-relational graph} for each layer of the student model (except for the prediction layer). The input $\mathbf{h}_m$ of the $m$-th graph is a set of node features containing all of the domain prototypes at $m$-th student layer, i.e., $\mathbf{h}_m = \{\mathbf{h}_{m,1}, ..., \mathbf{h}_{m,D}\} \in \mathcal{R}^{D \times F}$, where $D$ is the total domain number, $F$ is the channel number of each prototype.

In the first-layer domain-relational graph of the $m$-th student layer, a shared weight matrix $\mathbf{W}_m \in \mathcal{R}^{F^{\prime} \times F}$ is first applied to each node followed by a self-attention mechanism, where $F^{\prime}$ is the intermediate channel number. Then a multi-head concatenation mechanism with $K$ heads is employed to stabilize the training process. Specifically, each input prototype $\mathbf{h}_{m,d}$ is first transformed by the weight matrix $\mathbf{W}_m$, then the attention coefficient $\alpha_{i,j,m}$ between two nodes $i, j$ is calculated by applying a weight vector $\mathbf{a}_m \in \mathcal{R}^{2F^{\prime} \times 1}$ to the concatenation of their transformed features followed by the LeakyReLU nonlinearity and softmax function, which can be formulated as:
\begin{align}\label{eq: 5}
s_{i,j,m} &= \mathbf{a}_m^{\top}\left[\mathbf{W}_m \mathbf{h}_{m,i} \oplus \mathbf{W_m} \mathbf{h}_{m,j}\right], \\
\alpha_{i,j,m} &= \hbox{softmax}\left( \hbox{LeakyReLU}(s_{i,j,m}) \right) \nonumber \\
&= \frac{\exp \left(\hbox{LeakyReLU}(s_{i,j,m}) \right)}{\sum_{k \in \mathcal{N}_{i}} \exp \left(\hbox{LeakyReLU}(s_{i,k,m})\right)},
\end{align}
where $\oplus$ represents the concatenation operation and $\mathcal{N}_i$ is all the first-order neighbors of node $i$ (including node $i$). Then the final output $\mathbf{h}_{m,i}^{\prime} \in \mathcal{R}^{KF^{\prime}}$ of node $i$ can be obtained by the weighted sum of the transformed features of node $i$ and its neighbors based on their attention coefficients followed by the ELU nonlinearity and a multi-head concatenation mechanism:\begin{equation}\label{eq: h_i_prime}
{\mathbf{h}_{m,i}^{\prime}}=\oplus_{k=1}^{K} \hbox{ELU} (\sum_{j \in \mathcal{N}_{i}} \alpha_{i,j,m}^{k} \mathbf{W}_m^k \mathbf{h}_{m,j} ),
\end{equation} where $k$ represents the head index.

In the second-layer domain-relational graph of the $m$-th student layer, targeting at obtaining domain-relational ratios, we reformulate the parameters $\mathbf{W}_m, \mathbf{a}_m$ used in the first-layer graph as $\mathbf{W}_m^{\prime} \in \mathcal{R}^{1 \times KF^{\prime}}, \mathbf{a}_m^{\prime} \in \mathcal{R}^{2 \times 1}$ respectively and do not apply the multi-head mechanism. We use the softmax operation to normalize the output and finally derive the domain-relational ratios $r_m \in \mathcal{R}^D$, formulated as below:
\begin{align}
r_m = \hbox{softmax} ( \hbox{ELU} ( \sum_{j \in \mathcal{N}_i} \alpha_{i,j,m}^{\prime} \mathbf{W}_m^{\prime} \mathbf{h}_{m,j}^{\prime}) ),
\end{align}where $\alpha_{i,j,m}^{\prime}$ is calculated by:
\begin{align}\label{eq: 8}
s^{\prime}_{i,j,m} &= \mathbf{a}_m^{{\prime}^{\top}}\left[\mathbf{W}^{\prime}_m \mathbf{h}_{m,i}^{\prime} \oplus \mathbf{W^{\prime}_m} \mathbf{h}_{m,j}^{\prime} \right], \\
\alpha_{i,j,m}^{\prime} &= \hbox{softmax}\left( \hbox{LeakyReLU}(s^{\prime}_{i,j,m}) \right).
\end{align}

\subsection{Hierarchical Compare-aggregate Mechanism}\label{ssec: compare-aggregate}
As different domains might have different preferences towards different layer prototypes, we propose a \textit{hierarchical compare-aggregate mechanism} to dynamically select the most representative prototype for each domain. Our compare-aggregate mechanism is motivated by the Riesz representation theorem \cite{hartig1983the-riesz}, which indicates that an element can be evaluated by comparing it with a specific reference element and the quality of the element is the same as that of the selected reference element. Based on this, we establish a set of \textit{reference prototypes} for each domain and hierarchically aggregate the current and previous layer prototypes based on their similarities with the corresponding reference prototypes.

\paragraph{Reference prototype.} For each student layer, a simple way is to use the original domain prototypes of current layer as the reference prototypes for the current and previous layer prototypes. However, the information of other domains is not integrated, which plays an important role to enhance the model transferability across different domains. To handle this, we introduce a self-attention mechanism over all of the domain prototypes in the same layer to inject the information of different domains. Specifically, the reference prototype $\mathbf{RP}_m \in \mathcal{R}^{D \times F}$ of $m$-th student layer is calculated by:\begin{gather}
\mathbf{RP}_m = \boldsymbol{\alpha}_m^{\mathcal{D}} \cdot \mathbf{h}_m, \\
\boldsymbol{\alpha}_m^{\mathcal{D}} = \hbox{softmax}(\mathbf{h}_m \cdot \mathbf{W}_m^{\mathcal{D}} \cdot \mathbf{h}_m^{\top}),
\end{gather}where $\boldsymbol{\alpha}_m^{\mathcal{D}} \in \mathcal{R}^{D \times D}$ refers to the attention matrix of $m$-th layer, $\mathbf{h}_m \in \mathcal{R}^{D \times F}$ refers to the prototypes of all domains at $m$-th layer, $\mathbf{W}_m^{\mathcal{D}} \in \mathcal{R}^{F \times F}$ refers to a learnable parameter matrix at $m$-th layer, and the softmax operation is performed over the last vector dimension.

\paragraph{Compare-aggregate mechanism.} After obtaining the reference prototypes, we propose a compare-aggregate mechanism to hierarchically aggregate the layer prototypes by comparing them with the corresponding reference prototypes, which makes the model be aware to more representative layer prototypes for each domain. In detail, the aggregated prototype $\mathbf{AP}_{m,d} \in \mathcal{R}^{F}$ of $m$-th layer and $d$-th domain is formulated as:\begin{gather}
\mathbf{AP}_{m,d} = \boldsymbol{\alpha}_{m,d}^{\mathcal{H}} \cdot \mathbf{h}_{\leq m,d}, \\
\boldsymbol{\alpha}_{m,d}^{\mathcal{H}} = \hbox{softmax}(\mathbf{h}_{\leq m,d} \cdot \mathbf{W}_{m,d}^{\mathcal{H}} \cdot \mathbf{RP}_{m,d}),
\end{gather}where $\boldsymbol{\alpha}_{m,d}^{\mathcal{H}} \in \mathcal{R}^{m+1}$ represents the similarity ratios of $m$-th layer and $d$-th domain, $\mathbf{h}_{\leq m,d} \in \mathcal{R}^{(m+1) \times F}$ represents the prototypes of $m$-th layer and its previous layers at $d$-th domain, $\mathbf{W}_{m,d}^{\mathcal{H}} \in \mathcal{R}^{F \times F}$ is a learnable parameter matrix of $m$-th layer and $d$-th domain, and $\mathbf{RP}_{m,d} \in \mathcal{R}^{F}$ is the reference prototype of $m$-th layer and $d$-th domain. Then the aggregated prototype $\mathbf{AP}$ is sent to the domain-relational graphs to obtain the domain-relational ratios $r \in \mathcal{R}^{(M+1) \times D}$, as formulated by Equation (\ref{eq: 5})-(\ref{eq: 8}). 

Finally, the overall loss of our HRKD can be represented as:\begin{equation}
\resizebox{\columnwidth}{!}{
$\begin{aligned}
\mathcal{L}_{total} = \sum_{d=1}^D & \left( {r_{0,d} \mathcal{L}_{embd}^d + \sum_{m=1}^M r_{m,d} (\mathcal{L}_{attn}^{m,d} + \mathcal{L}_{hidn}^{m,d})} \right. \\ 
& \left. {+ \frac{\gamma}{D} \mathcal{L}_{pred}^d} \right),
\end{aligned}$}
\end{equation}where $r_{m,d}$ is the domain-relational ratio at $m$-th student layer and $d$-th domain.

\section{Experiment}
In this section, we conduct extensive experiments on two multi-domain datasets, namely MNLI and Amazon Reviews, to demonstrate the effectiveness of our HRKD method.

\subsection{Datasets and Model Settings}
We evaluate our method on two multi-domain datasets, including the multi-genre natural language inference (MNLI) dataset \citep{williams2018a-broad-converage-mnli} and the Amazon Reviews dataset \citep{blitzer2007biographies-amazon-review}. In detail, MNLI is a natural language inference dataset with five domains for the task of entailment relation prediction between two sentences. In our setting, we randomly sample 10\% of the original training data as our development set and use the original development set as our test set. Amazon Reviews is a sentiment analysis dataset with four domains for predicting whether the reviews are positive or negative. 
Following \citet{pan2020meta-kd}, we randomly split the original data into train, development, and test sets. The statistics of these two datasets are listed in Table \ref{tab: dataset statistics}.
\begin{table}
\centering
\caption{Statistics of the MNLI and Reviews datasets.}\label{tab: dataset statistics}
\resizebox{\columnwidth}{!}{
\begin{tabular}[b]{c c c c c}
\toprule
\textbf{Dataset}& \textbf{Domain}& \textbf{\#Train}& \textbf{\#Dev}& \textbf{\#Test} \\
\midrule
\multirow{5}{*}{MNLI}& Fiction& 69,613& 7,735& 1,973 \\
~& Government& 69,615& 7,735& 1,945 \\
~& Slate& 69,575& 7,731& 1,955 \\
~& Telephone& 75,013& 8,335& 1,966 \\
~& Travel& 69,615& 7,735& 1,976 \\
\midrule
\multirow{4}{2cm}{\centering Amazon Reviews}& Books& 1,631& 170& 199 \\
~& DVD& 1,621& 194& 185 \\
~& Electronics& 1,615& 172& 213 \\
~& Kitchen& 1,613& 184& 203 \\
\bottomrule
\end{tabular}}
\end{table}

We use $\hbox{BERT}_{\rm B}$ (the number of layers $N$=12, the hidden size $d^{\prime}$=768, the FFN intermediate hidden size $d_i^{\prime}$=3072, the number of attention heads $h$=12, the number of parameters \#params=109M) as the architecture of our teacher model, and $\hbox{BERT}_{\rm S}$ ($M$=4, $d^{\prime}$=312, $d_i^{\prime}$=1200, $h$=12, \#params=14.5M) as our student model. Our teacher model $\hbox{HRKD-teacher}$ is trained in a multi-domain manner as described in Section \ref{ssec: multi-domain kd}, and our student model $\hbox{BERT}_{\rm S}$ is initialized with the general distillation weights of TinyBERT\footnote{We use the 2nd version from \url{https://github.com/huawei-noah/Pretrained-Language-Model/tree/master/TinyBERT}}.

\begin{table*}[th!]
\centering
\caption{Results on MNLI in terms of accuracy (\%) with standard deviations. $X \xrightarrow{\scriptsize \hbox{A}} Y$ denotes using teacher $X$ to distill student $Y$ with KD method of $A$. The bold and underlined numbers indicate the best and the second-best performance, respectively.}\label{tab: mnli results}
\resizebox{0.95\textwidth}{!}{
\begin{tabular}{l ccccc c}
\toprule
\textbf{Method}& \textbf{Fiction}& \textbf{Government}& \textbf{Slate}& \textbf{Telephone}& \textbf{Travel}& \textbf{Average} \\
\midrule
$\hbox{BERT}_{\rm B}$-single& 82.2& 84.2& 76.7& 82.4& 84.2& 81.9 \\
$\hbox{BERT}_{\rm B}$-mix& 84.8& 87.2& 80.5& 83.8& 85.5& 84.4 \\
$\hbox{BERT}_{\rm B}$-mtl& 83.7& 87.1& 80.6& 83.9& 85.8& 84.2 \\
$\hbox{Meta-teacher}$& 85.1& 86.5& 81.0& 83.9& 85.5& 84.4 \\
$\hbox{HRKD-teacher}$& 83.8& 87.6& 80.4& 83.5& 85.4& 84.2 \\
\midrule
$\hbox{BERT}_{\rm B}$-single $\xrightarrow{\scriptsize \hbox{TinyBERT-KD}}$ $\hbox{BERT}_{\rm S}$& 78.8& 83.2& 73.6& 78.8& 81.9& 79.3 \\
$\hbox{BERT}_{\rm B}$-mix $\xrightarrow{\scriptsize \hbox{TinyBERT-KD}}$ $\hbox{BERT}_{\rm S}$& 79.6& 83.3& 74.8& 79.0& 81.5& 79.6 \\
$\hbox{BERT}_{\rm B}$-mtl $\xrightarrow{\scriptsize \hbox{TinyBERT-KD}}$ $\hbox{BERT}_{\rm S}$& 79.7& 83.1& 74.2& 79.3& 82.0& 79.7 \\
$\hbox{Meta-teacher}$ $\xrightarrow{\scriptsize \hbox{Meta-distillation}}$ $\hbox{BERT}_{\rm S}$& \textbf{80.5}& 83.7& 75.0& \uline{80.5}& \uline{82.1}& \uline{80.4} \\
\midrule
$\hbox{HRKD-teacher}$ $\xrightarrow{\scriptsize \hbox{TinyBERT-KD}}$ $\hbox{BERT}_{\rm S}$& 80.1$\pm$0.22& \uline{84.2$\pm$0.20}& \uline{75.7$\pm$0.27}& 80.0$\pm$0.23& 81.9$\pm$0.17& \uline{80.4} \\
$\hbox{HRKD-teacher}$ $\xrightarrow{\scriptsize \hbox{HRKD}}$ $\hbox{BERT}_{\rm S}$& \uline{80.4$\pm$0.33}& \textbf{84.3}$\pm$0.30& \textbf{76.1}$\pm$0.32& \textbf{81.4}$\pm$0.29& \textbf{82.2}$\pm$0.26& \textbf{80.9} \\
\bottomrule
\end{tabular}
}
\end{table*}
\begin{table*}[th!]
\centering
\caption{Results on Amazon Reviews in terms of accuracy (\%) with standard deviations.}\label{tab: amazon results}
\resizebox{0.83\textwidth}{!}{
\begin{tabular}{l cccc c}
\toprule
\textbf{Method}& \textbf{Books}& \textbf{DVD}& \textbf{Electronics}& \textbf{Kitchen}& \textbf{Average} \\
\midrule
$\hbox{BERT}_{\rm B}$-single& 87.9& 83.8& 89.2& 90.6& 87.9 \\
$\hbox{BERT}_{\rm B}$-mix& 89.9& 85.9& 90.1& 92.1& 89.5 \\
$\hbox{BERT}_{\rm B}$-mtl& 90.5& 86.5& 91.1& 91.1& 89.8 \\
$\hbox{Meta-teacher}$& 92.5& 87.0& 91.1& 89.2& 89.9 \\
$\hbox{HRKD-teacher}$& 88.4& 89.2& 92.5& 91.1& 90.3 \\
\midrule
$\hbox{BERT}_{\rm B}$-single $\xrightarrow{\scriptsize \hbox{TinyBERT-KD}}$ $\hbox{BERT}_{\rm S}$& 83.4& 83.2& 89.2& \uline{91.1}& 86.7 \\
$\hbox{BERT}_{\rm B}$-mix $\xrightarrow{\scriptsize \hbox{TinyBERT-KD}}$ $\hbox{BERT}_{\rm S}$& 88.4& 81.6& 89.7& 89.7& 87.3 \\
$\hbox{BERT}_{\rm B}$-mtl $\xrightarrow{\scriptsize \hbox{TinyBERT-KD}}$ $\hbox{BERT}_{\rm S}$& \uline{90.5}& 81.6& 88.7& 90.1& 87.7 \\
$\hbox{Meta-teacher}$ $\xrightarrow{\scriptsize \hbox{Meta-distillation}}$ $\hbox{BERT}_{\rm S}$& \textbf{91.5}& 86.5& 90.1& 89.7& \uline{89.4} \\
\midrule
$\hbox{HRKD-teacher}$ $\xrightarrow{\scriptsize \hbox{TinyBERT-KD}}$ $\hbox{BERT}_{\rm S}$& 84.6$\pm$0.93& \uline{87.8$\pm$0.55}& \uline{91.3$\pm$0.23}& 88.1$\pm$2.98& 87.9 \\
$\hbox{HRKD-teacher}$ $\xrightarrow{\scriptsize \hbox{HRKD}}$ $\hbox{BERT}_{\rm S}$& 87.4$\pm$0.90& \textbf{90.5}$\pm$1.76& \textbf{91.8}$\pm$1.25& \textbf{92.2}$\pm$0.48& \textbf{90.5} \\
\bottomrule
\end{tabular}
}
\end{table*}

\subsection{Baselines}
We mainly compare our KD method with several KD baseline methods distilled from four teacher models, including $\hbox{BERT}_{\rm B}$-single, $\hbox{BERT}_{\rm B}$-mix, $\hbox{BERT}_{\rm B}$-mtl, and Meta-teacher in Meta-KD \citep{pan2020meta-kd}. Specifically, $\hbox{BERT}_{\rm B}$-single trains the teacher model of each domain separately with the single-domain dataset; $\hbox{BERT}_{\rm B}$-mix trains a single teacher model with the combined dataset of all domains; $\hbox{BERT}_{\rm B}$-mtl adopts the multi-task training method proposed by \citet{liu2019multi-task} to train the teacher model; Meta-teacher trains the teacher model with several meta-learning strategies including prototype-based instance weighting and domain corruption.

\subsection{Implementation Details}
For the teacher model, we train the HRKD-teacher for three epochs with a learning rate of 5e-5. For the student model, we train it for ten epochs with a learning rate of 1e-3 and 5e-4 on MNLI and Amazon Reviews, respectively. $\gamma$ is set to 1, and $t$ is 1. For few-shot learning, the learning rate for the student model is 5e-5, while other hyper-parameters are kept the same. {The few-shot training data is selected from the front of our original training set with different sample ratios, while the dev and test data are the same as our original dev and test sets without sampling to make a fair comparison.} In all the experiments, the sequence length is set to 128, and the batch size is 32. The hyper-parameters are tuned on the development set, and the results are averaged over five runs. Our experiments are conducted on 4 GeForce RTX 3090 GPUs.

\subsection{General Experimental Results}
The experimental results of our method are shown in Table \ref{tab: mnli results} and \ref{tab: amazon results}. On the MNLI dataset, our teacher model HRKD-teacher has similar performances with other baseline teacher models, but the performance of the student model distilled with the HRKD method ($\hbox{HRKD-teacher}$ $\xrightarrow{\scriptsize \hbox{HRKD}}$ $\hbox{BERT}_{\rm S}$) is significantly better than the base TinyBERT-KD method ($\hbox{HRKD-teacher}$ $\xrightarrow{\scriptsize \hbox{TinyBERT-KD}}$ $\hbox{BERT}_{\rm S}$) as well as its counterpart Meta-KD ($\hbox{Meta-teacher}$ $\xrightarrow{\scriptsize \hbox{Meta-distillation}}$ $\hbox{BERT}_{\rm S}$), which demonstrate the superior performance of our method. Specifically, with the HRKD method, the average score of the student model is both 0.5\% higher than that of the model with the base TinyBERT-KD method and its counterpart Meta-KD method (see Table \ref{tab: mnli results}). It can also be observed that the improvement of our HRKD method on the Telephone domain is the most significant, which is probably caused by the amount of training data. From Table \ref{tab: dataset statistics}, we can see that the Telephone domain has much more training data than other domains, indicating that the Telephone domain can derive more relationship information from other domains and lead to higher improvement. Meanwhile, as shown in the results on the Amazon Reviews dataset in Table \ref{tab: amazon results}, the performance of the HRKD-teacher model is slightly better than that of other teacher models, but the student model distilled by the HRKD method largely outperforms the models distilled by the TinyBERT-KD and Meta-KD methods with average gains of 2.6\% and 1.1\% respectively, which prove the excellent performance of our method again. Note that our HRKD method significantly outperforms the base TinyBERT-KD method on both MNLI and Amazon Reviews datasets (t-test with p < 0.1). And since the performances of the Meta-teacher and our HRKD-teacher are similar on both datasets, the impact of the teacher is negligible, making the comparison between our HRKD and its counterpart Meta-KD relatively fair.

\begin{figure}
\centering
\includegraphics[width=0.8\columnwidth]{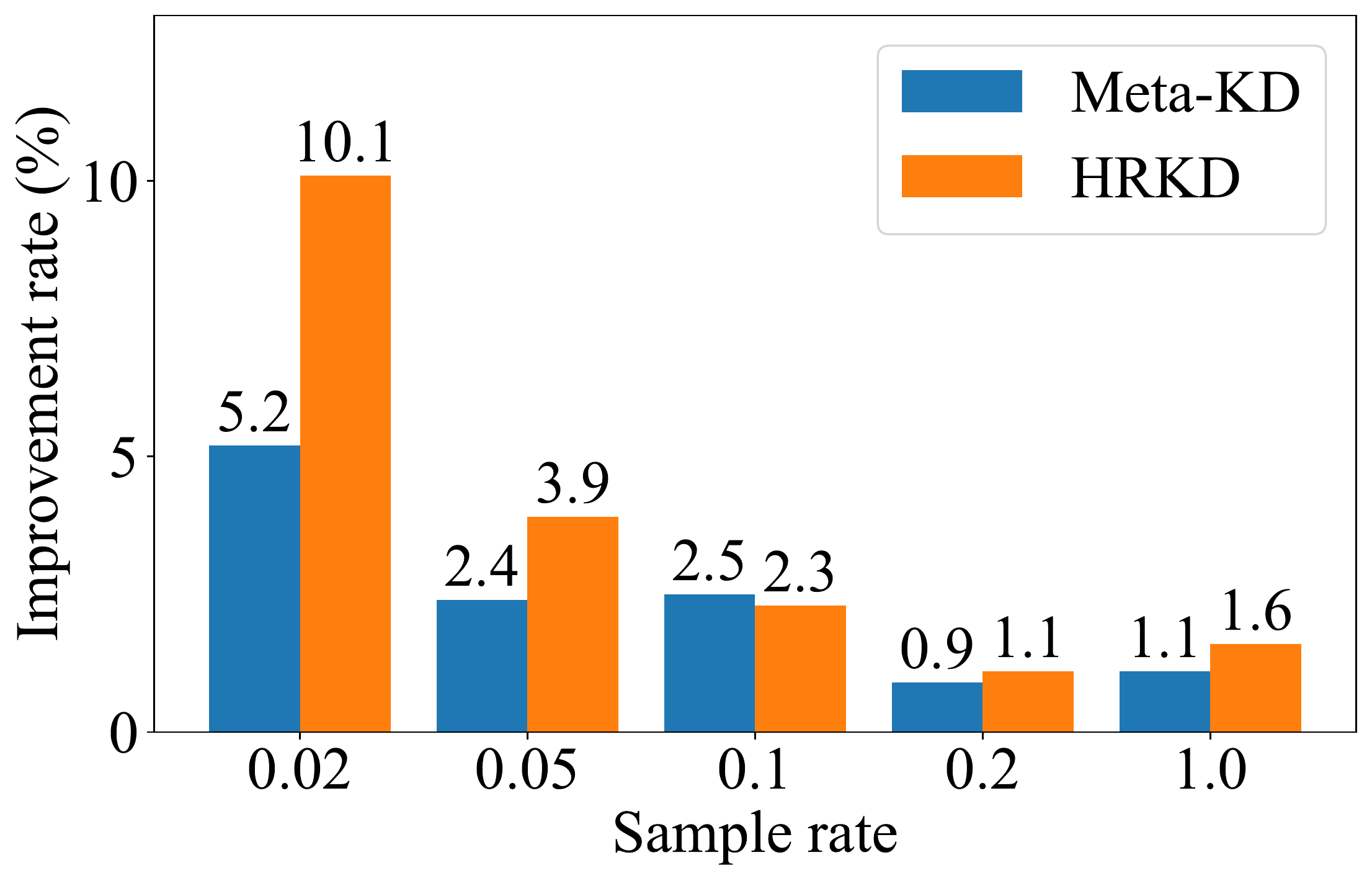}
\caption{Comparison results of few-shot learning between Meta-KD and HRKD.}\label{fig: few shot}
\end{figure}

\subsection{Few-shot Learning Results}
As a large amount of training data is hard to collect in reality, the few-shot learning ability of our method is worth being evaluated, where both the teacher and student models are trained with few training data in each domain. We randomly sample a part of the training data in the MNLI dataset to make an evaluation, where the chosen sample rates are 2\%, 5\%, 10\%, and 20\%. We mainly compare the performance improvements between two methods: distilling from $\hbox{BERT}_{\rm B}$-single to $\hbox{BERT}_{\rm S}$ with TinyBERT-KD ($\hbox{BERT}_{\rm B}$-single $\xrightarrow{\scriptsize \hbox{TinyBERT-KD}}$ $\hbox{BERT}_{\rm S}$) and our HRKD method ($\hbox{HRKD-teacher}$ $\xrightarrow{\scriptsize \hbox{HRKD}}$ $\hbox{BERT}_{\rm S}$). From the results in Figure \ref{fig: few shot}, we can observe that the improvement gets more prominent when the training data gets fewer, and the average improvement rate is the largest of 10.1\% when there is only 2\% MNLI training data. In addition, we can see that the improvement rates of our method are higher than those of Meta-KD under most of the sample rates, especially when there are only 2\% training data. These results demonstrate the strong learning ability of our HRKD method under the few-shot setting.

\begin{table}
\centering
\caption{Ablation studies on Amazon Reviews dataset. The accuracy values (\%) are reported.}\label{tab: kd ablation results}
\resizebox{\columnwidth}{!}{
\begin{tabular}{l cccc c}
\toprule
\textbf{KD Method}& \textbf{Books}& \textbf{DVD}& \textbf{Elec.}& \textbf{Kitchen}& \textbf{Average} \\
\midrule
HRKD& \textbf{87.4} & 90.5 & 91.8 & \textbf{92.2} & \textbf{90.5} \\
\midrule
- Self-attention& 87.3 & 89.9 & \textbf{92.5} & 91.6 & 90.3 \\
- Comp-Agg& 86.3 & \textbf{90.6} & 90.9 & 91.8 & 89.9 \\
- Hierarchical Rel.& 85.9 & 89.6 & 91.1 & 91.6 & 89.5 \\
- Domain Rel.& 84.6 & 87.8 & 91.3 & 88.1 & 87.9 \\
\bottomrule
\end{tabular}}
\end{table}

\begin{table*}[th!]
\centering
\caption{Case study on Amazon Reviews across four domains with three positive samples and one negative sample. Positive samples are colored in gray.}\label{tab: case study 1}
\resizebox{0.95\textwidth}{!}{
\begin{tabular}{c|c|c|c|c}
\toprule
\textbf{Domain}& \textbf{Label}& \multicolumn{3}{c}{\textbf{Review Text}} \\
\midrule
\rowcolor{Gray}
Books& POS& \multicolumn{3}{c}{...leading, or molding young people today would benefit from reading this book...} \\
\rowcolor{Gray}
DVD& POS& \multicolumn{3}{c}{...The plot wasn't horrible, it was actually pretty good for a fright flick...} \\
Electronics& NEG& \multicolumn{3}{c}{...I returned the camera and bought a Panasonic and never looked back!} \\
\rowcolor{Gray}
Kitchen& POS& \multicolumn{3}{c}{This is great for making poached eggs on toast. My family has enjoyed using it...} \\
\midrule
\multicolumn{2}{c|}{\textbf{Domain-relational Ratio}}& \multicolumn{3}{c}{\textbf{Hierarchical Similarity Ratio}} \\
\midrule
\rowcolor{Gray}
\multicolumn{2}{c|}{[0.24, 0.29, 0.25, 0.23, 0.24]}& \multicolumn{3}{c}{[0.42, 0.58], [0.38, 0.30, 0.32], [0.24, 0.27, 0.25, 0.24], [0.16, 0.19, 0.21, 0.20, 0.24]} \\
\rowcolor{Gray}
\multicolumn{2}{c|}{[0.24, 0.29, 0.25, 0.23, 0.24]}& \multicolumn{3}{c}{[0.52, 0.48], [0.32, 0.38, 0.30], [0.27, 0.24, 0.27, 0.22], [0.16, 0.17, 0.19, 0.24, 0.24]} \\
\multicolumn{2}{c|}{[0.27, 0.17, 0.25, 0.25, 0.26]}& \multicolumn{3}{c}{[0.62, 0.38], [0.22, 0.21, 0.58], [0.19, 0.22, 0.30, 0.29], [0.14, 0.15, 0.24, 0.21, 0.26]} \\
\rowcolor{Gray}
\multicolumn{2}{c|}{[0.24, 0.25, 0.25, 0.28, 0.27]}& \multicolumn{3}{c}{[0.46, 0.54], [0.30, 0.43, 0.28], [0.30, 0.25, 0.26, 0.20], [0.29, 0.19, 0.15, 0.12, 0.26]} \\
\bottomrule
\end{tabular}}
\end{table*}
\begin{table*}[th!]
\centering
\caption{Case study on Amazon Reviews across four domains with two positive samples and two negative samples. Positive samples are colored in gray.}\label{tab: case study 2}
\resizebox{0.95\textwidth}{!}{
\begin{tabular}{c|c|c|c|c}
\toprule
\textbf{Domain}& \textbf{Label}& \multicolumn{3}{c}{\textbf{Review Text}} \\
\midrule
Books& NEG& \multicolumn{3}{c}{...In this book, his "hard-evidence" is flimsey and suspicious...} \\
\rowcolor{Gray}
DVD& POS& \multicolumn{3}{c}{...If you have a child who loves John Deere, then this is a perfect DVD for them.} \\
\rowcolor{Gray}
Electronics& POS& \multicolumn{3}{c}{...Movies are amazing! My music collection never sounded so good...} \\
Kitchen& NEG& \multicolumn{3}{c}{This is the worst blender I've ever used...It's also loud and it moves a lot...} \\
\midrule
\multicolumn{2}{c|}{\textbf{Domain-relational Ratio}}& \multicolumn{3}{c}{\textbf{Hierarchical Similarity Ratio}} \\
\midrule
\multicolumn{2}{c|}{[0.26, 0.25, 0.25, 0.24, 0.27]}& \multicolumn{3}{c}{[0.43, 0.57], [0.40, 0.31, 0.29], [0.25, 0.26, 0.25, 0.23], [0.13, 0.17, 0.20, 0.21, 0.30]} \\
\rowcolor{Gray}
\multicolumn{2}{c|}{[0.24, 0.25, 0.25, 0.26, 0.22]}& \multicolumn{3}{c}{[0.52, 0.48], [0.35, 0.32, 0.33], [0.22, 0.25, 0.27, 0.25], [0.18, 0.22, 0.18, 0.20, 0.22]} \\
\rowcolor{Gray}
\multicolumn{2}{c|}{[0.24, 0.25, 0.25, 0.26, 0.25]}& \multicolumn{3}{c}{[0.54, 0.46], [0.31, 0.33, 0.36], [0.22, 0.24, 0.29, 0.25], [0.21, 0.20, 0.18, 0.18, 0.23]} \\
\multicolumn{2}{c|}{[0.26, 0.25, 0.25, 0.24, 0.27]}& \multicolumn{3}{c}{[0.49, 0.51], [0.36, 0.33, 0.31], [0.26, 0.25, 0.26, 0.23], [0.22, 0.20, 0.19, 0.18, 0.21]} \\
\bottomrule
\end{tabular}}
\end{table*}

\subsection{Ablation Studies}
In this section, we progressively remove each module of our KD method to evaluate the effect of each module. 

The results are shown in Table \ref{tab: kd ablation results}. We first remove the self-attention mechanism across different domain prototypes (- Self-attention), and the average score on Amazon Reviews drops by 0.2\%, which proves its effectiveness. Next, we replace the hierarchical compare-aggregate mechanism with a simple average operation (- Comp-Agg), and the average score drops by 0.4\%, which demonstrates the effectiveness of the compare-aggregate mechanism. {Then we remove the hierarchical graph structure (- Hierarchical Rel.), where the input of each domain-relational graph comes from a single student layer. As can be seen, the average score drops by 0.4\%, which proves the importance of the hierarchical relationship. Finally, we remove the domain-relational graph in each layer (- Domain Rel.), and the performance significantly drops by 1.6\%, which strongly demonstrates the advantage of the domain relationship.}

\subsection{Case Studies}
We further provide some case studies to intuitively explain the effectiveness of the domain-relational ratios and hierarchical similarity ratios calculated by our HRKD method (see Table \ref{tab: case study 1} and \ref{tab: case study 2}). 

In Table \ref{tab: case study 1} and \ref{tab: case study 2}, we use the label to denote the categories of sampled domain examples, and we assume that if the learned domain-relational ratios and hierarchical similarity ratios are similar for domain examples with same category while different for those with different categories, then the model has relatively correctly captured the cross-domain and hierarchical relational information. We select two typical types of cases from Amazon Reviews across four domains, in which we adjust the number of domains in each category under two settings: (i) three same categories (i.e., POS) with another one category (i.e., NEG) as in Table \ref{tab: case study 1}, and (ii) two same categories (i.e., POS) with another two same categories (i.e., NEG) as in Table \ref{tab: case study 2}. 

We find the results are intuitive, as we observe that the review texts with the same labels have similar domain-relational ratios and hierarchical similarity ratios, while different layers indeed have different domain weighting preferences and different preferences of layer prototypes for graph input. For example, in Table \ref{tab: case study 1} and \ref{tab: case study 2}, positive samples tend to have higher domain-relational ratios in the middle layers (i.e., 2-4), while negative samples have higher ratios in the marginal layers (i.e., 1, 5). Meanwhile, in the second and third layers of Table \ref{tab: case study 1} as well as the first layer of Table \ref{tab: case study 2}, lower positive layer prototypes tend to have higher similarity ratios, and the higher positive layer prototypes in the third layer of Table \ref{tab: case study 2} also tend to have higher similarity ratios; while those of the negative layer prototypes are just the opposite. The results show that HRKD method has distinctively and correctly captured the hierarchical and domain meta-knowledges, leading to better performance.

\section{Related Work}
\paragraph{Pre-trained Language Model (PLM) Compression.} Due to the large size and slow inference speed, PLMs are hard to be deployed on edge devices for practical usage. To solve this problem, many PLM compression methods have been proposed, including quantization \cite{shen2020q-bert}, weight pruning \cite{michel2019are-sixteen}, and knowledge distillation (KD) \cite{sun2019patient,jiao2020tinybert}. Among them, KD \citep{hinton2015distilling} has been widely adopted due to its plug-and-play feasibility, aiming to distill the knowledge from a larger teacher model to a smaller student model without decreasing too much performance. For example, BERT-PKD \citep{sun2019patient} distills both intermediate and output layers on fine-tuning. TinyBERT \citep{jiao2020tinybert} additionally distills the embedding layer and attention matrices during pre-training and fine-tuning. Meta-KD \citep{pan2020meta-kd} proposes to distill knowledge from a cross-domain meta-teacher through an instance-specific domain-expertise weighting technique. 

In this paper, we propose a novel cross-domain KD framework that captures the relational information across different domains with both domain and hierarchical meta-knowledges, which has a better capability for capturing multi-domain correlations. 

\paragraph{Transfer Learning and Meta-learning.} Transfer learning focuses on transferring the knowledge from source domains to boost the model performance on the target domain. Among the methods in transfer learning, the shared-private architecture \citep{liu2017adversarial,liu2019multi-task} is most commonly applied in NLP tasks, which consists of a shared network to store domain-invariant knowledge and a private network to capture domain-specific information. There are also many works applying adversarial training strategies \cite{shen2018wasserstein,li2019transferable,zhou2019dual}, which introduce domain adversarial classifiers to learn the domain-invariant features. Besides, the research of multi-domain learning has gained more and more attention recently, which is a particular case of transfer learning targeting transferring knowledge across different domains to comprehensively enhance the model performance \citep{cai2019multi-domain,wang2020meta}. Unlike transfer learning, the goal of meta-learning is to train a meta-learner that can easily adapt to a new task with a few training data and iterations \citep{finn2017model-agnostic}. Traditional meta-learning typically contains three categories of methods: metric-based \citep{snell2017prototypical,sung2018learning}, model-based \citep{santoro2016meta-learning,munkhdalai2017meta}, and optimization-based \citep{ravi2017optimization,finn2017model-agnostic}. In addition, the meta-learning technique can benefit the multi-domain learning task by learning the relationship information among different domains \citep{franceschi2017forward}. 

In this paper, we leverage meta-learning to solve the multi-domain learning task, where we consider cross-domain KD to simultaneously capture the correlation between different domains, aiming to train a better student meta-learner.

\section{Conclusion}
In this paper, we present a hierarchical relational knowledge distillation (HRKD) framework to simultaneously capture the cross-domain relational information. We build several domain-relational graphs to capture domain meta-knowledge and introduce a hierarchical compare-aggregate mechanism to capture hierarchical meta-knowledge. The learnt domain-relational ratios are leveraged to measure domain importance during the KD process. Extensive experiments on public datasets demonstrate the superior performance and solid few-shot learning ability of our HRKD method.

\section*{Acknowledgements}
This work was financially supported by the National Natural Science Foundation of China (No.61602013).

\normalem
\bibliography{emnlp2021}

\begin{thebibliography}{37}
\expandafter\ifx\csname natexlab\endcsname\relax\def\natexlab#1{#1}\fi

\bibitem[{Bao et~al.(2020)Bao, He, Wang, Wu, and Wang}]{bao2020plato}
Siqi Bao, Huang He, Fan Wang, Hua Wu, and Haifeng Wang. 2020.
\newblock {PLATO}: Pre-trained dialogue generation model with discrete latent
  variable.
\newblock In \emph{Proceedings of the 58th Annual Meeting of the Association
  for Computational Linguistics}, pages 85--96.

\bibitem[{Blitzer et~al.(2007)Blitzer, Dredze, and
  Pereira}]{blitzer2007biographies-amazon-review}
John Blitzer, Mark Dredze, and Fernando Pereira. 2007.
\newblock Biographies, {B}ollywood, boom-boxes and blenders: Domain adaptation
  for sentiment classification.
\newblock In \emph{Proceedings of the 45th Annual Meeting of the Association of
  Computational Linguistics}, pages 440--447.

\bibitem[{Cai and Wan(2019)}]{cai2019multi-domain}
Yitao Cai and Xiaojun Wan. 2019.
\newblock Multi-domain sentiment classification based on domain-aware embedding
  and attention.
\newblock In \emph{Proceedings of the Twenty-Eighth International Joint
  Conference on Artificial Intelligence}, pages 4904--4910.

\bibitem[{CONNEAU and Lample(2019)}]{conneau2019cross-lingual}
Alexis CONNEAU and Guillaume Lample. 2019.
\newblock Cross-lingual language model pretraining.
\newblock In \emph{Advances in Neural Information Processing Systems}.

\bibitem[{Devlin et~al.(2019)Devlin, Chang, Lee, and
  Toutanova}]{devlin2019bert}
Jacob Devlin, Ming-Wei Chang, Kenton Lee, and Kristina Toutanova. 2019.
\newblock {BERT}: Pre-training of deep bidirectional transformers for language
  understanding.
\newblock In \emph{Proceedings of the 2019 Conference of the North {A}merican
  Chapter of the Association for Computational Linguistics: Human Language
  Technologies}, pages 4171--4186.

\bibitem[{Finn et~al.(2017)Finn, Abbeel, and Levine}]{finn2017model-agnostic}
Chelsea Finn, Pieter Abbeel, and Sergey Levine. 2017.
\newblock Model-agnostic meta-learning for fast adaptation of deep networks.
\newblock In \emph{Proceedings of the 34th International Conference on Machine
  Learning}, pages 1126--1135.

\bibitem[{Franceschi et~al.(2017)Franceschi, Donini, Frasconi, and
  Pontil}]{franceschi2017forward}
Luca Franceschi, Michele Donini, Paolo Frasconi, and Massimiliano Pontil. 2017.
\newblock Forward and reverse gradient-based hyperparameter optimization.
\newblock In \emph{Proceedings of the 34th International Conference on Machine
  Learning}, pages 1165--1173.

\bibitem[{Hartig(1983)}]{hartig1983the-riesz}
Donald~G. Hartig. 1983.
\newblock The riesz representation theorem revisited.
\newblock \emph{The American Mathematical Monthly}, pages 277--280.

\bibitem[{Hinton et~al.(2015)Hinton, Vinyals, and Dean}]{hinton2015distilling}
Geoffrey Hinton, Oriol Vinyals, and Jeffrey Dean. 2015.
\newblock Distilling the knowledge in a neural network.
\newblock In \emph{NIPS Deep Learning and Representation Learning Workshop}.

\bibitem[{Javed and White(2019)}]{javed2019meta-learning}
Khurram Javed and Martha White. 2019.
\newblock Meta-learning representations for continual learning.
\newblock In \emph{Advances in Neural Information Processing Systems}.

\bibitem[{Jiao et~al.(2020)Jiao, Yin, Shang, Jiang, Chen, Li, Wang, and
  Liu}]{jiao2020tinybert}
Xiaoqi Jiao, Yichun Yin, Lifeng Shang, Xin Jiang, Xiao Chen, Linlin Li, Fang
  Wang, and Qun Liu. 2020.
\newblock {T}iny{BERT}: Distilling {BERT} for natural language understanding.
\newblock In \emph{Findings of the Association for Computational Linguistics:
  EMNLP 2020}, pages 4163--4174.

\bibitem[{Li et~al.(2019)Li, Li, Wei, Bing, Zhang, and
  Yang}]{li2019transferable}
Zheng Li, Xin Li, Ying Wei, Lidong Bing, Yu~Zhang, and Qiang Yang. 2019.
\newblock Transferable end-to-end aspect-based sentiment analysis with
  selective adversarial learning.
\newblock In \emph{Proceedings of the 2019 Conference on Empirical Methods in
  Natural Language Processing and the 9th International Joint Conference on
  Natural Language Processing}, pages 4590--4600.

\bibitem[{Liu et~al.(2019{\natexlab{a}})Liu, Wang, Lin, Socher, and
  Xiong}]{liu2019mkd}
Linqing Liu, Huan Wang, Jimmy Lin, Richard Socher, and Caiming Xiong.
  2019{\natexlab{a}}.
\newblock Mkd: a multi-task knowledge distillation approach for pretrained
  language models.
\newblock \emph{arXiv preprint arXiv:1911.03588}.

\bibitem[{Liu et~al.(2017)Liu, Qiu, and Huang}]{liu2017adversarial}
Pengfei Liu, Xipeng Qiu, and Xuanjing Huang. 2017.
\newblock Adversarial multi-task learning for text classification.
\newblock In \emph{Proceedings of the 55th Annual Meeting of the Association
  for Computational Linguistics}, pages 1--10.

\bibitem[{Liu et~al.(2019{\natexlab{b}})Liu, He, Chen, and
  Gao}]{liu2019improving}
Xiaodong Liu, Pengcheng He, Weizhu Chen, and Jianfeng Gao. 2019{\natexlab{b}}.
\newblock Improving multi-task deep neural networks via knowledge distillation
  for natural language understanding.
\newblock \emph{arXiv preprint arXiv:1904.09482}.

\bibitem[{Liu et~al.(2019{\natexlab{c}})Liu, He, Chen, and
  Gao}]{liu2019multi-task}
Xiaodong Liu, Pengcheng He, Weizhu Chen, and Jianfeng Gao. 2019{\natexlab{c}}.
\newblock Multi-task deep neural networks for natural language understanding.
\newblock In \emph{Proceedings of the 57th Annual Meeting of the Association
  for Computational Linguistics}, pages 4487--4496.

\bibitem[{Liu and Lapata(2019)}]{liu2019text}
Yang Liu and Mirella Lapata. 2019.
\newblock Text summarization with pretrained encoders.
\newblock In \emph{Proceedings of the 2019 Conference on Empirical Methods in
  Natural Language Processing and the 9th International Joint Conference on
  Natural Language Processing}, pages 3730--3740.

\bibitem[{Lu et~al.(2015)Lu, Behbood, Hao, Zuo, Xue, and
  Zhang}]{lu2015transfer}
Jie Lu, Vahid Behbood, Peng Hao, Hua Zuo, Shan Xue, and Guangquan Zhang. 2015.
\newblock Transfer learning using computational intelligence: A survey.
\newblock \emph{Knowledge-Based Systems}, pages 14--23.

\bibitem[{Michel et~al.(2019)Michel, Levy, and Neubig}]{michel2019are-sixteen}
Paul Michel, Omer Levy, and Graham Neubig. 2019.
\newblock Are sixteen heads really better than one?
\newblock In \emph{Advances in Neural Information Processing Systems}, pages
  14014--14024.

\bibitem[{Munkhdalai and Yu(2017)}]{munkhdalai2017meta}
Tsendsuren Munkhdalai and Hong Yu. 2017.
\newblock Meta networks.
\newblock In \emph{Proceedings of the 34th International Conference on Machine
  Learning}, pages 2554--2563.

\bibitem[{Pan et~al.(2020)Pan, Wang, Qiu, Zhang, Li, and
  Huang}]{pan2020meta-kd}
Haojie Pan, Chengyu Wang, Minghui Qiu, Yichang Zhang, Yaliang Li, and Jun
  Huang. 2020.
\newblock Meta-kd: A meta knowledge distillation framework for language model
  compression across domains.
\newblock \emph{arXiv preprint arXiv:2012.01266}.

\bibitem[{Peng et~al.(2020)Peng, Ji, Lin, Cui, Chen, and Zhang}]{peng2020mtss}
Shuke Peng, Feng Ji, Zehao Lin, Shaobo Cui, Haiqing Chen, and Yin Zhang. 2020.
\newblock Mtss: Learn from multiple domain teachers and become a multi-domain
  dialogue expert.
\newblock In \emph{Proceedings of the AAAI Conference on Artificial
  Intelligence}, pages 8608--8615.

\bibitem[{Ravi and Larochelle(2017)}]{ravi2017optimization}
Sachin Ravi and Hugo Larochelle. 2017.
\newblock Optimization as a model for few-shot learning.
\newblock In \emph{International Conference on Learning Representations}.

\bibitem[{Santoro et~al.(2016)Santoro, Bartunov, Botvinick, Wierstra, and
  Lillicrap}]{santoro2016meta-learning}
Adam Santoro, Sergey Bartunov, Matthew Botvinick, Daan Wierstra, and Timothy
  Lillicrap. 2016.
\newblock Meta-learning with memory-augmented neural networks.
\newblock In \emph{Proceedings of The 33rd International Conference on Machine
  Learning}, pages 1842--1850.

\bibitem[{Shen et~al.(2018)Shen, Qu, Zhang, and Yu}]{shen2018wasserstein}
Jian Shen, Yanru Qu, Weinan Zhang, and Yong Yu. 2018.
\newblock Wasserstein distance guided representation learning for domain
  adaptation.
\newblock In \emph{Proceedings of the Thirty-Second {AAAI} Conference on
  Artificial Intelligence}, pages 4058--4065.

\bibitem[{Shen et~al.(2020)Shen, Dong, Ye, Ma, Yao, Gholami, Mahoney, and
  Keutzer}]{shen2020q-bert}
Sheng Shen, Zhen Dong, Jiayu Ye, Linjian Ma, Zhewei Yao, Amir Gholami,
  Michael~W. Mahoney, and Kurt Keutzer. 2020.
\newblock Q-bert: Hessian based ultra low precision quantization of bert.
\newblock \emph{Proceedings of the AAAI Conference on Artificial Intelligence},
  pages 8815--8821.

\bibitem[{Snell et~al.(2017)Snell, Swersky, and Zemel}]{snell2017prototypical}
Jake Snell, Kevin Swersky, and Richard Zemel. 2017.
\newblock Prototypical networks for few-shot learning.
\newblock In \emph{Advances in Neural Information Processing Systems}.

\bibitem[{Sun et~al.(2019)Sun, Cheng, Gan, and Liu}]{sun2019patient}
Siqi Sun, Yu~Cheng, Zhe Gan, and Jingjing Liu. 2019.
\newblock Patient knowledge distillation for {BERT} model compression.
\newblock In \emph{Proceedings of the 2019 Conference on Empirical Methods in
  Natural Language Processing and the 9th International Joint Conference on
  Natural Language Processing}, pages 4323--4332.

\bibitem[{Sung et~al.(2018)Sung, Yang, Zhang, Xiang, Torr, and
  Hospedales}]{sung2018learning}
Flood Sung, Yongxin Yang, Li~Zhang, Tao Xiang, Philip~H.S. Torr, and Timothy~M.
  Hospedales. 2018.
\newblock Learning to compare: Relation network for few-shot learning.
\newblock In \emph{IEEE/CVF Conference on Computer Vision and Pattern
  Recognition}, pages 1199--1208.

\bibitem[{Veličković et~al.(2018)Veličković, Cucurull, Casanova, Romero,
  Liò, and Bengio}]{velickovic2018graph}
Petar Veličković, Guillem Cucurull, Arantxa Casanova, Adriana Romero, Pietro
  Liò, and Yoshua Bengio. 2018.
\newblock Graph attention networks.
\newblock In \emph{International Conference on Learning Representations}.

\bibitem[{Wang et~al.(2020)Wang, Qiu, Huang, and He}]{wang2020meta}
Chengyu Wang, Minghui Qiu, Jun Huang, and Xiaofeng He. 2020.
\newblock Meta fine-tuning neural language models for multi-domain text mining.
\newblock In \emph{Proceedings of the 2020 Conference on Empirical Methods in
  Natural Language Processing (EMNLP)}, pages 3094--3104.

\bibitem[{Williams et~al.(2018)Williams, Nangia, and
  Bowman}]{williams2018a-broad-converage-mnli}
Adina Williams, Nikita Nangia, and Samuel Bowman. 2018.
\newblock A broad-coverage challenge corpus for sentence understanding through
  inference.
\newblock In \emph{Proceedings of the 2018 Conference of the North {A}merican
  Chapter of the Association for Computational Linguistics: Human Language
  Technologies}, pages 1112--1122.

\bibitem[{Yang et~al.(2020)Yang, Shou, Gong, Lin, and Jiang}]{yang2020model}
Ze~Yang, Linjun Shou, Ming Gong, Wutao Lin, and Daxin Jiang. 2020.
\newblock Model compression with two-stage multi-teacher knowledge distillation
  for web question answering system.
\newblock In \emph{Proceedings of the 13th International Conference on Web
  Search and Data Mining}, page 690–698.

\bibitem[{Zhang et~al.(2019)Zhang, Wei, and Zhou}]{zhang2019hibert}
Xingxing Zhang, Furu Wei, and Ming Zhou. 2019.
\newblock {HIBERT}: Document level pre-training of hierarchical bidirectional
  transformers for document summarization.
\newblock In \emph{Proceedings of the 57th Annual Meeting of the Association
  for Computational Linguistics}, pages 5059--5069.

\bibitem[{Zheng et~al.(2020)Zheng, Zhang, Huang, and
  Mao}]{zheng2020a-pre-training}
Yinhe Zheng, Rongsheng Zhang, Minlie Huang, and Xiaoxi Mao. 2020.
\newblock A pre-training based personalized dialogue generation model with
  persona-sparse data.
\newblock \emph{Proceedings of the AAAI Conference on Artificial Intelligence},
  pages 9693--9700.

\bibitem[{Zhou et~al.(2019)Zhou, Zhang, Jin, Zhu, Fang, Goh, and
  Kwok}]{zhou2019dual}
Joey~Tianyi Zhou, Hao Zhang, Di~Jin, Hongyuan Zhu, Meng Fang, Rick Siow~Mong
  Goh, and Kenneth Kwok. 2019.
\newblock Dual adversarial neural transfer for low-resource named entity
  recognition.
\newblock In \emph{Proceedings of the 57th Annual Meeting of the Association
  for Computational Linguistics}, pages 3461--3471.

\bibitem[{Zhu et~al.(2020)Zhu, Xia, Wu, He, Qin, Zhou, Li, and
  Liu}]{zhu2020incorporating}
Jinhua Zhu, Yingce Xia, Lijun Wu, Di~He, Tao Qin, Wengang Zhou, Houqiang Li,
  and Tieyan Liu. 2020.
\newblock Incorporating bert into neural machine translation.
\newblock In \emph{International Conference on Learning Representations}.

\end{thebibliography}
\bibliographystyle{acl_natbib}




\end{document}